\documentclass{article}
\usepackage{spconf}

\usepackage{spconf,amsmath,graphicx,hyperref}
\usepackage{amsfonts}
\usepackage{array}
\usepackage[caption=false,font=normalsize,labelfont=sf,textfont=sf]{subfig}
\usepackage{textcomp}
\usepackage{stfloats}
\usepackage{url}
\usepackage{verbatim}
\usepackage{balance}
\usepackage{xcolor}

\usepackage[english]{babel}
\usepackage{bm,bbm,amsmath,amssymb,amsthm,graphicx}

\usepackage{url}
\usepackage{algorithm}
\usepackage{algpseudocode}
\usepackage{pifont}
\usepackage{todonotes}
\usepackage{tikz}
\usepackage{pdfsync}
\usepackage{float}      
\usepackage{placeins} 
\usepackage{caption}
\usepackage{graphicx}
\usepackage{subcaption} 
\usetikzlibrary{calc,patterns,decorations.pathmorphing,decorations.markings}

\usepackage{comment}
\usepackage[normalem]{ulem}

\newcommand{\MAT}{ [ \begin{array}}  
\newcommand{\mat}{\end{array}  ]}

\newtheorem{Q}{Question}[section]

\newcommand{\argmin}{\operatorname{argmin}}

\newcommand{\vct}[1]{\mathbf{#1}}
\newcommand{\tTx}[1]{\mathbf{#1}}

\newcommand{\bmtx}[1]{\mathbf{#1}}

\newcommand{\calC}{\mathcal{C}}

\newcommand{\ve}{\vct{e}}

\newcommand{\vk}{\vct{k}}

\newcommand{\vo}{\vct{o}}

\newcommand{\vq}{\vct{q}}

\newcommand{\vv}{\vct{v}}

\newcommand{\vx}{\vct{x}}
\newcommand{\vy}{\vct{y}}
\newcommand{\vz}{\vct{z}}

\newcommand{\vxi}{\vct{\xi}}

\newcommand{\mH}{\tTx{H}}

\newcommand{\mS}{\tTx{S}}

\newcommand{\mW}{\tTx{W}}

\newcommand{\mId}{\bmtx{I}}

\newcommand{\mc}{\mathcal}

\usepackage{cleveref}
\usepackage[square,numbers,compress]{natbib}

\title{In-Context Learning for Non-Stationary MIMO Equalization}
\name{Jiachen Jiang$^{1}$, Zhen Qin$^{2,3,4}$, Zhihui Zhu$^{1}$
\thanks{This work was supported in part by NSF grants IIS-2312840, IIS-2402952, and ECCS-240971. ZQ gratefully acknowledges support from the MICDE Research Scholars Program at the University of Michigan.}}
\address{ $^{1}$ Department of Computer Science and Engineering, The Ohio State University, Columbus, USA \\
$^{2}$ Michigan Institute for Computational Discovery and Engineering, University of Michigan, Ann Arbor, USA \\
$^{3}$ Department of Electrical Engineering and Computer Science, University of Michigan, Ann Arbor, USA \\
$^{4}$ Department of Statistics, University of Michigan, Ann Arbor, USA}

\begin{document}
\maketitle

\begin{abstract}
Channel equalization is fundamental for mitigating distortions such as frequency-selective fading and inter-symbol interference. Unlike standard supervised learning approaches that require costly retraining or fine-tuning for each new task, in-context learning (ICL) adapts to new channels at inference time with only a few examples. However, existing ICL-based equalizers are primarily developed for and evaluated on static channels within the context window. Indeed, to our knowledge, prior principled analyses and theoretical studies of ICL focus exclusively on the stationary setting, where the function remains fixed within the context. In this paper, we investigate the ability of ICL to address non-stationary problems through the lens of time-varying channel equalization. We employ a principled framework for designing efficient attention mechanisms with improved adaptivity in non-stationary tasks, leveraging algorithms from adaptive signal processing to guide better designs. For example, new attention variants can be derived from the Least Mean Square (LMS) adaptive algorithm, a Least Root Mean Square (LRMS) formulation for enhanced robustness, or multi-step gradient updates for improved long-term tracking. Experimental results demonstrate that ICL holds strong promise for non-stationary MIMO equalization, and that attention mechanisms inspired by classical adaptive algorithms can substantially enhance adaptability and performance in dynamic environments. Our findings may provide critical insights for developing next-generation wireless foundation models with stronger adaptability and robustness.

\end{abstract}
\begin{keywords}
Channel equalization, in-context learning, test-time regression, time-varying channels, LMS/LRMS.
\end{keywords}

\section{Introduction}
\label{sec:intro}

Channel equalization in communication systems is a fundamental technique that compensates for distortions introduced by the channel, such as frequency-selective fading and inter-symbol interference, thereby improving the reliability and throughput of data transmission. Numerous approaches have been developed for channel equalization, including classical methods such as zero-forcing (ZF) equalization \citep{tuchler2002minimum}, linear minimum mean square error (LMMSE) equalizers \citep{bashar2020uplink}, adaptive equalizers \citep{qureshi2005adaptive}, decision feedback equalization (DFE) \citep{belfiore1979decision}, and blind equalizers \citep{benveniste2003blind}. In addition, recent advances in machine learning have enabled data-driven equalizers, such as deep learning-based \citep{ye2017initial}, meta-learning-based \citep{zhang2022meta}, deep reinforcement learning-based \citep{choi2023deep}, and graph neural network-based equalizers \citep{clausius2024graph}, which can capture complex channel impairments and adapt to varying propagation conditions.

Despite these advances, previous learning-based equalizers typically require training a separate model for each task in conventional single-task learning frameworks or involve additional parameter updates and optimization loops in joint-learning approaches such as meta-learning and hypernetworks. Recently, a new paradigm---in-context learning (ICL)---has emerged \citep{min2021metaicl}. ICL allows models to perform previously unseen tasks at inference time by conditioning on sequences of input-output examples, without necessitating any explicit parameter updates. This capability has shown promising results when applied to channel equalization \citep{zecchin2024context,zecchin2024cell,song2024context,abbas2024leveraging}, enabling the model to rapidly adapt to new channel realizations using only a few examples.

However, in the context of time-varying channels \citep{tang2007pilot,chiong2015channel,qin2018time,li2019time,tao2019sparse,wang2022sparse,qin2023dynamic}, the channel matrix evolves dynamically and is often only partially observable. Consequently, equalizers must continuously adapt based on limited or noisy observations, rendering the equalization problem a high-dimensional and temporally dependent inference task. Existing ICL-based equalizers, however, typically assume a static channel within the context window and employ standard softmax attention, which may hinder their ability to capture rapid channel variations and temporal correlations. To our knowledge, prior work on principled analyses or theoretical studies of in-context learning follow the framework of learning a function class under the assumption that the function remains fixed within the context window~\citep{garg2022can,yang2024context,jiang2025compression}. This motivates us to investigate the following questions:
\begin{center}
\textit{Can in-context learning not only identify the task from the context but also track time-varying variations of the task?\\ Is softmax attention the optimal choice in non-stationary settings, or can new variants of attention mechanisms be developed to enhance adaptivity?}
\end{center}

In this paper, we study these questions in the context of time-varying channels. First, we propose a new framework that extends ICL from a function class to a class of time-varying functions, and we instantiate it for the problem of channel equalization. To develop new variants of attention mechanisms, we draw inspiration from recent work on the connection between transformers and Kalman filters~\citep{goel2024can}, test-time regression~\citep{wang2025test}, and state-space models. Building on this, we employ a principled framework for designing efficient attention mechanisms with improved adaptivity, leveraging algorithms from adaptive signal processing to guide improved designs. Applied to channel equalization, the Least Mean Square (LMS) algorithm—a widely used adaptive filtering method—naturally induces the DeltaNet~\citep{yang2024parallelizing} attention. In addition, we propose a multi-step update strategy to capture longer-term dynamics and further improve adaptation. We further employ the Least Root Mean Square (LRMS) formulation to design attention mechanisms with enhanced robustness.  Experimental results demonstrate that in-context learning is promising for non-stationary multiple-
input multiple-output (MIMO)  equalization, and that new attention variants inspired by classical adaptive algorithms can substantially enhance adaptability and performance in dynamic environments.

\section{Preliminaries}
\label{sec:preli}

\subsection{Channel Model}
Since wireless channels exhibit temporal correlation, it is essential to employ a model that captures both short-term variations and long-term dependencies. To this end, we consider the time-varying $m_1 \times m_2$ MIMO autoregressive model \citep{li2019time,hijazi2010channel,prasad2015joint,wu2021channel}:

\begin{equation}
    \mH_{i} = \rho \mH_{i-1} + \sqrt{1 - \rho^2} \mW_{i}, \quad i = 2, \ldots, K,
\label{eq:channel_model}
\end{equation}
where $\rho \in [0,1)$ denotes the \textit{memory factor}, which controls the time-varying rate of the channel, $\mH_{i} \in \mathbb{C}^{m_2 \times m_1}$ represents the complex-valued channel matrix and   $\mW_{i}\sim\mc C \mc N(\mathbf{0}, \sigma_{w}^2 \mathbf{I})$ is the additive noise matrix.  

Building on the time-varying channel model in \eqref{eq:channel_model}, we consider a discrete-time MIMO system with additive noise and finite-resolution quantization, modeled as
\begin{equation}
    \vy_{i} = Q_b(\mH_{i} \vx_{i} + \ve_{i}), \quad i = 1,\ldots,K,
\label{eq:channel}
\end{equation}
where $\vx_i \in \mathbb{C}^{m_1}$ and $\vy_i \in \mathbb{C}^{m_2}$ denote the transmitted and received signals, respectively; $\ve_i$ is additive Gaussian noise with variance $\sigma^2$, and $Q_b(\cdot)$ is a $b$-bit quantizer. The transmitted signal is normalized such that $\mathbb{E}[\|\vx_{i}\|^2] = 1$, which ensures an average signal-to-noise ratio (SNR) of $\mathrm{SNR} = \mathbb{E}[\|\vx_{i}\|^2] /  \sigma^2 = 1/\sigma^2$.


\subsection{LMMSE Equalization}
To perform channel equalization, the goal is to recover the transmitted signals $\vx_i \in \mathbb{C}^{m_1}$ from the received data $\vy_i \in \mathbb{C}^{m_2}$.
A commonly used approach is the Linear Minimum Mean Square Error (LMMSE) estimator, which computes a linear mapping $\mW$ that minimizes the mean square error between the transmitted signal and its linear estimate from the noisy observation: $
    \mathbb{E}\!\left[ \left\| \vx_{i} - \mW\vy_{i} \right\|_{2}^{2} \right].$

Suppose that the transmitted signal $\vx_i$ follows a zero-mean Gaussian distribution, the quantizer is neglected, and the channel matrix $\mH_{K+1}$ is perfectly known. Under these assumptions, the LMMSE estimate of the input signal given the received observation $\vy_{K+1}$ is
\begin{equation}
    \hat{\vx}_{\mathrm{LMMSE}}^{*}
= \bigl( 2\sigma^{2} I + {\mH}_{K+1}^{\mathrm{H}} {\mH}_{K+1} \bigr)^{-1} {\mH}_{K+1}^{\mathrm{H}} \vy_{K+1}.
\label{eq:LMMSE}
\end{equation}
This estimator provides a well-defined baseline for comparison, as it assumes perfect channel knowledge and circumvents the challenges associated with time-varying channel estimation.

\subsection{ICL-Based Equalization}
While the LMMSE estimator provides a well-defined baseline, it relies on knowledge of the instantaneous channel matrix and does not directly leverage observed input-output pairs to adapt to time-varying dynamics. In contrast, ICL-based equalization aims to exploit a set of previous input-output pairs (context $\calC = \{(\vx_i, \vy_i)\}_{i=1}^{K}$) to infer the transmitted signal $\vx_{K+1}$ from a new received observation $\vy_{K+1}$ without explicit knowledge of the underlying channel.
The equalizer, parameterized by $\theta$, produces an estimate of $\vx_{K+1}$ as
\begin{equation}
\hat{\vx}_{K+1} = f_{\theta}(\mathcal{C}, \vy_{K+1}).
\end{equation}
Its performance is quantified by the mean squared error (MSE) between the estimated signal and the ground truth: \begin{equation*}
    \mathrm{MSE}(\theta) 
= \mathbb{E}_{\{(\vx_{i}, \vy_{i})_{i=1}^{K+1}\}} 
\left[ \left\| f_{\theta}(\mc C, \vy_{K+1}) - \vx_{K+1} \right\|^2 \right].
\label{eq:MSE}
\end{equation*}
Here, the expectation is taken over the distribution of input-output pairs. A lower MSE indicates better equalization performance. For time-varying channels, capturing the dynamics from the context is more challenging, which necessitates a more robust ICL-based equalizer.



\section{Adaptive Inference for ICL-Based Equalization}
\label{sec:method}

Softmax attention is widely used in transformers to model dependencies in sequential inputs. 
For a sequence of vectors $\vz_1, \dots, \vz_K\in\mathbb{R}^{d}$, attention computes a weighted sum of past observations where the weights are determined by pairwise similarities between queries and keys. 
Formally, each input vector is first projected into query, key, and value vectors, and the output of the attention layer is obtained as:
\begin{equation}
\begin{aligned}
    \vq_i, \vk_i, \vv_i &= \mW_{Q} \vz_i, \mW_{K} \vz_i, \mW_{V} \vz_i,  \\
    \vo_i &= \sum_{j = 1}^{i} \frac{\mathrm{exp}(\vk_{j}^{\top}\vq_{i})}{\sum_{k = 1}^{i} \mathrm{exp}(\vk_{k}^{\top}\vq_{i})} \vv_j =: \xi_i(\vq_i),
\end{aligned}
\label{eq:softmax-attention}\end{equation}
where $\mW_{Q}, \mW_{K}, \mW_{V} \in \mathbb{R}^{d \times d}$ are learnable projection matrices for the query $\vq_i$, key $\vk_i$, and value $\vv_i$. Under the ICL setting, each input vector $\vz_i$ corresponds to a transmitted or received signal, i.e., $\vx_i$ or $\vy_i$, and the resulting attention output $\vo_i \in \mathbb{R}^{d}$ captures contextual information from previous observations.

However, prior work~\citep{liu2022non} demonstrates that softmax attention is sensitive to non-stationarity, leading to degraded performance on time-varying sequences. Motivated by this limitation, in this section we develop adaptive attention mechanisms that can effectively track temporal dynamics and improve performance in non-stationary MIMO equalization tasks.

\subsection{Least Mean Square (LMS) Attention}
\label{subsubsec:attn_framework}
To motivate the design, note that when the softmax function is removed in \eqref{eq:softmax-attention}, the output becomes $\vo_i = \vxi_i (\vq_i) = (\sum_{j=1}^i \vv_j \vk_j^\top) \vq_i$, where $(\sum_{j=1}^i \vv_j \vk_j^\top)$ stores the information of previous tokens. To better capture the time-varying property  of the input-output caused by the non-stationary channel \eqref{eq:channel_model},  
we use a state space model $\vo_i = \mS_i \vq_i$ with a learnable state matrix $\mS_i \in \mathbb{R}^{d \times d}$ that serves to accumulate and propagate contextual information over time by (approximately) solving the following test-time regression~\citep{wang2025test}:
\begin{equation}
    \mS_i \approx \argmin_{\mS\in\mathbb{R}^{d\times d}} \; \mc L(\mS) = \frac{1}{2} \sum_{j=1}^i\| \vv_j - \mS \vk_j\|_2^2.\label{eq:attn_framework}
\end{equation}
By updating $\mS_i$ during inference, the model effectively adapts to temporal variations in the channel and captures timely information. Inspired by the classical Least Mean Squares (LMS) algorithm in adaptive filtering, we perform a one-step gradient-descent update of $\mS_i$ for the current observation:
\begin{equation}
        \mS_i 
= \mS_{i-1} - \beta_i \big( \mS_{i-1}\vk_i - \vv_i \big)\vk_i^{\top},
\label{eq:LMS_attn}
\end{equation}
where $\beta_i$ is a learnable parameter that controls the history writing strength. The LMS attention 
can be equivalently rewritten as 
$\mS_i = \mS_{i-1}(\mId - \beta_i\vk_i\vk_i^{\top}) + \beta_i\vv_i\vk_i^{\top}$ which corresponds to a structure in the DeltaNet attention, where the state is updated by combining a retained past component $\mS_{i-1}(\mId - \beta_i\vk_i\vk_i^{\top})$ with a new information component $\beta_i\vv_i\vk_i^{\top}$. This demonstrates that the LMS-inspired update provides a principled connection between classical adaptive filtering and the DeltaNet attention mechanism \citep{wang2025test}.

\subsection{Multi-Step LMS (Multi-LMS) Attention}
\label{subsubsec:Multi-GD_attn}
While the single-step LMS update provides a simple mechanism for updating $\mS_i$, 
it may not quickly adapt to the variations of the channel matrix, resulting in degraded long-term accuracy. 
To improve the adaptation speed and stability, we introduce an $M$-step closed-form extension of LMS to solve \eqref{eq:attn_framework}, which leads to the following multi-step LMS attention:
\begin{equation}
    \mS_{i} = \mS_{i-1} - \frac{1- (1-\beta_i\|\vk_i\|_2^2)^{M}}{\|\vk_i\|_2^2} (\mS_{i-1} \vk_i - \vv_i) \vk_i^{\top}.
\end{equation}
\begin{proof}
Set $\mS_{i}^{(0)} = \mS_{i-1}$ as the initialization and $\mS_i = \mS_i^{(M)}$ as the final update. 
According to \eqref{eq:LMS_attn}, the $m$-th update of $\mS$ can be written as
\begin{eqnarray}\mS_{i}^{(m+1)}
= \mS_i^{(m)} - \beta_i\,\ve_i^{(m)}\vk_i^{\!\top}
\label{iterative equation of error S}
\end{eqnarray}
with $\ve_{i}^{(m)} = \mS_{i}^{(m)}\vk_i - \vv_i$. Substituting \eqref{iterative equation of error S} into the error recursion yields 
\begin{eqnarray}
       \ve_{i}^{(m+1)} &\!\!\!\!=\!\!\!\!& \mS_{i}^{(m+1)}\vk_i - \vv_i = (1 - \beta_i\|\vk_i\|_2^2)\ve_{i}^{(m)}\nonumber\\
       &\!\!\!\!=\!\!\!\!& (1 - \beta_i\|\vk_i\|_2^2)^{(m+1)}\ve_{i}^{(0)}.
\label{iterative equation of error e}
\end{eqnarray}
Furthermore, we can derive
\begin{eqnarray}
       \mS_{i} &\!\!\!\!\triangleq\!\!\!\!&\mS_{i}^{(M)}= \mS_{i}^{(0)} -\beta_i\sum_{m=0}^{M-1}(1 - \beta_i\|\vk_i\|_2^2)^{(m)}\ve_{i}^{(0)}\vk_i^\top \nonumber\\
       &\!\!\!\!=\!\!\!\!& \mS_{i}^{(0)} - \frac{1- (1-\beta_i\|\vk_i\|_2^2)^{M}}{\|\vk_i\|_2^2} \ve_{i}^{(0)} \vk_i^{\top}.
\label{final M step closed form}
\end{eqnarray}

\end{proof}

\subsection{Least Root Mean Square (LRMS) Attention}
\label{subsubsec:LRMS_attn}

The MSE loss in \ref{eq:attn_framework} is statistically optimal when the noise follows a Gaussian distribution, owing to its convexity and smoothness. However, its quadratic growth with respect to the residual makes it highly sensitive to outliers and less robust in dynamic or non-stationary channel environments. To alleviate these limitations, we adopt the root-mean-square (RMS) loss, which increases only linearly with the residual magnitude and therefore reduces the impact of occasional large errors. This formulation yields a more robust objective under practical conditions:
\begin{equation}
    \mS_i \approx \argmin_{\mS\in\mathbb{R}^{d\times d}} \; \mc L(\mS) = \frac{1}{2} \sum_{j=1}^i\| \vv_j - \mS \vk_j\|_2.
    \label{eq:RMS_loss}
\end{equation}
Following the derivation of LMS attention, we now present the recurrent form of Least Root Mean Square (LRMS) attention:
\begin{eqnarray}
       \mS_{i} = \mS_{i-1} - \beta_i \frac{(\mS_{i-1} \vk_i - \vv_i)}{\|\mS_{i-1} \vk_i - \vv_i \|_2} \vk_i^{\top}.
\label{The recurrent form of LRMS}
\end{eqnarray}
Compared to \Cref{eq:LMS_attn}, the LRMS attention introducing an additional normalization term on the prediction residual, allowing the model to adaptively down-weight unreliable updates and stabilize learning.

\section{Numerical Experiment}
\label{sec:exp}

\begin{figure*}[t]
    \centering
    \subfloat[Memory Factor]
    {\includegraphics[width=0.32\textwidth]{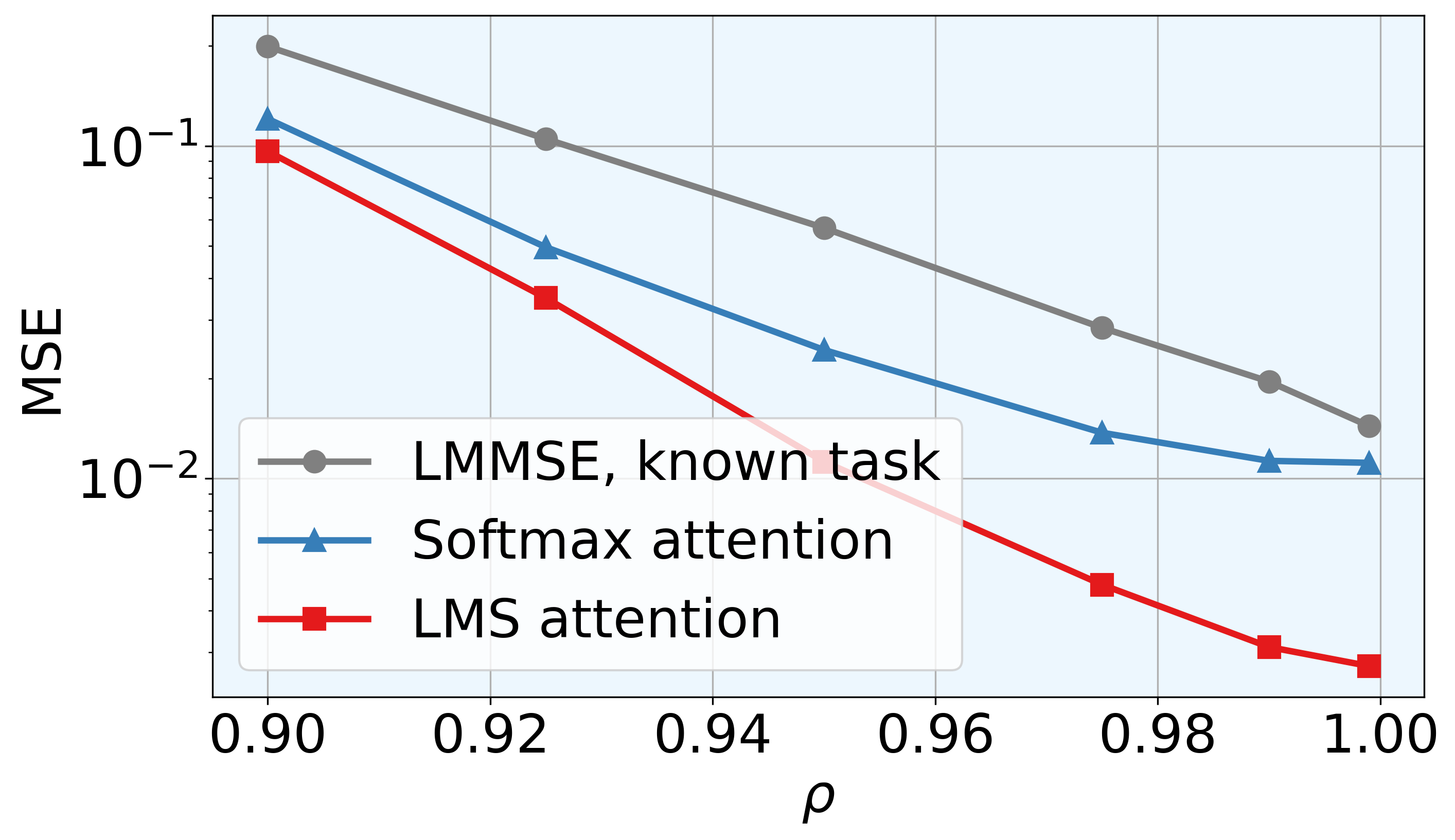}\label{fig:rho}}
    \hfill
    \subfloat[\text{SNR}]
    {\includegraphics[width=0.32\textwidth]{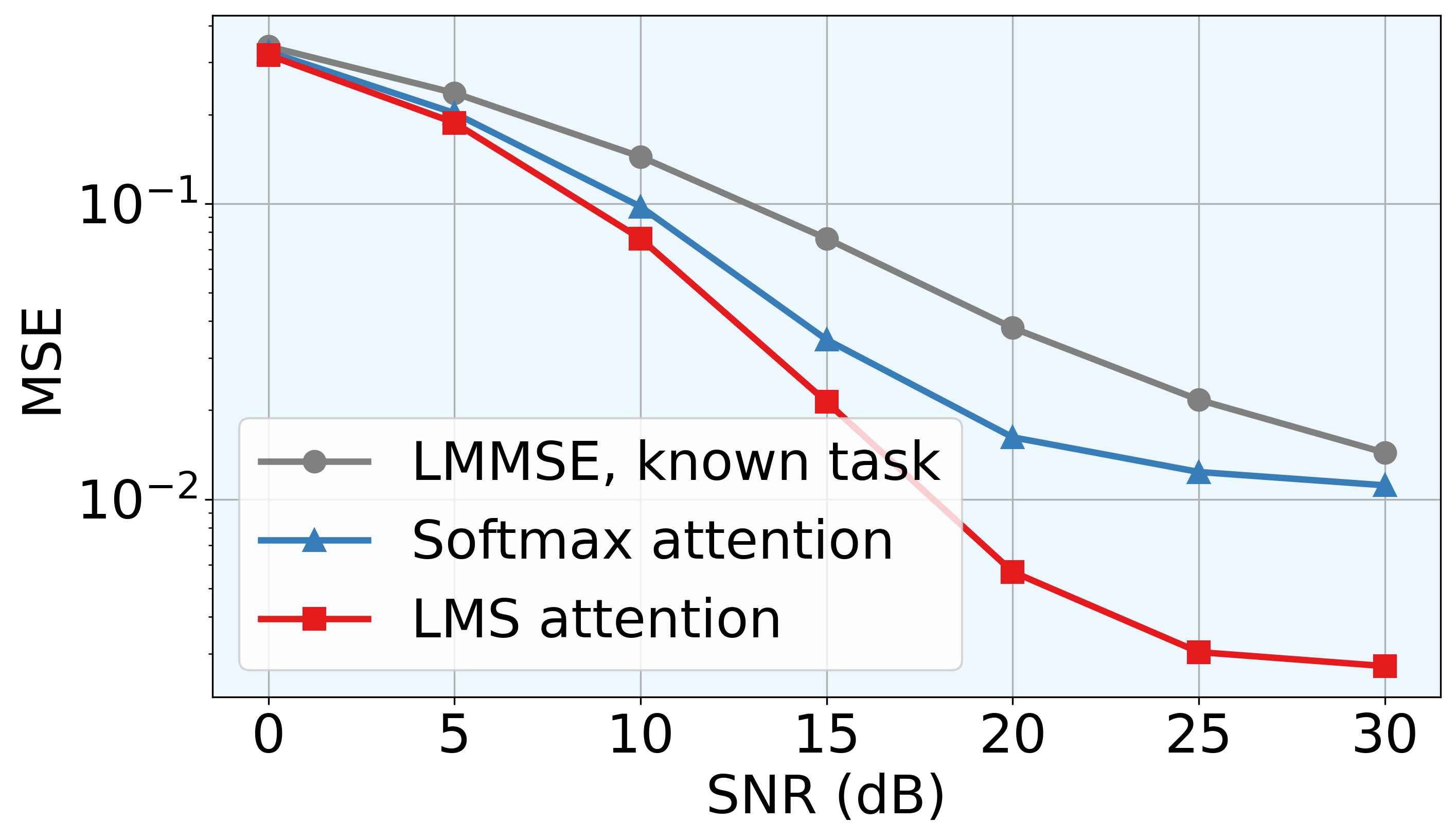}\label{fig:SNR}}
    \hfill
    \subfloat[Quantization Bits]
    {\includegraphics[width=0.32\textwidth]{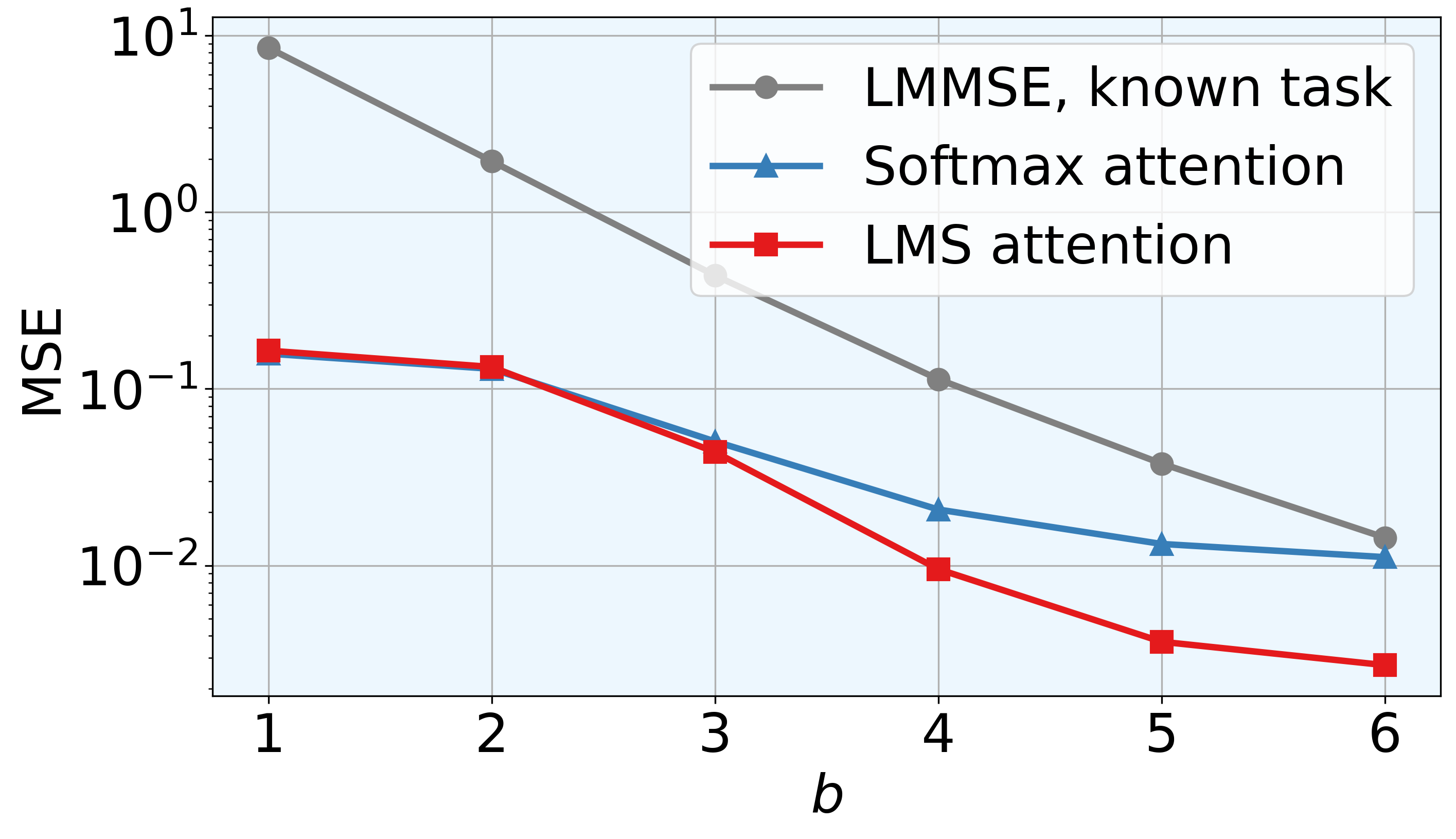}\label{fig:bits}}
    \caption{Performance comparison of the LMMSE with two ICL-based equalizers with softmax attention and LMS attention across (a) memory factors $\rho \in [0.9, 1.0]$; (b) $\mathrm{SNR} \in [0, 30]$ dB, and (c) quantization bits $b\in[1,6]$. }
    \vspace{-.4in}
    \label{fig:rho_snr_bits}
\end{figure*}

In our numerical experiments, similar to \citep{zecchin2024context}, we train a two-layer GPT-2 transformer (embedding dimension 64, 4 attention heads per layer) on a synthetic $2 \times 2$ time-varying MIMO dataset. For the training set, the channel input $\vx$ is drawn from a normalized QPSK constellation and the received signal is quantized by a $b$-bit uniform quantizer with range $[-4,4]$. The initial channel matrix $\mH_1$ is sampled from the complex Gaussian distribution $\mathcal{CN}(0,\mathbf{I})$. The channel-variation noise level is set as $\sigma_w = 0.1$. During training, the memory factor $\rho$ is uniformly sampled from $[0.9, 1)$, the per-channel SNR from $[0, 30]$ dB, and the quantization bits $b$ from the integer range $[1, 6]$. Each sequence has context length $K=20$. The total number of pretraining channels is $8192$. Models are trained with MSE loss using the Adam optimizer for $50000$ steps with batch size of $128$. For validation, we generate $1000$ independent time-varying channel instances, each with fixed values of $\rho$, SNR, and $b$. Unless otherwise specified, we set $\rho = 0.99$, $\mathrm{SNR} = 30$ dB, and $b = 6$.


In the first experiment, we compare the performance of LMMSE and ICL-based equalizers using both softmax and LMS attention, as shown in \Cref{fig:rho,fig:SNR,fig:bits}. Across all figures, increasing the memory factor $\rho$, SNR, or quantization bits consistently reduces the estimation error. Notably, the ICL-based equalizers outperform LMMSE under all settings, and LMS attention exhibits comparable or superior performance relative to softmax attention. Moreover, unlike softmax attention, which incurs a longer training time due to the computational overhead of the nonlinear function, LMS attention not only reduces this computational burden but also maintains or improves accuracy, highlighting its practical advantage.

\begin{figure}[t]
  \centering  
  \subfloat[Multi- vs. Single-Step\label{fig:robust_m_step}]{
    \includegraphics[width=0.45\linewidth]{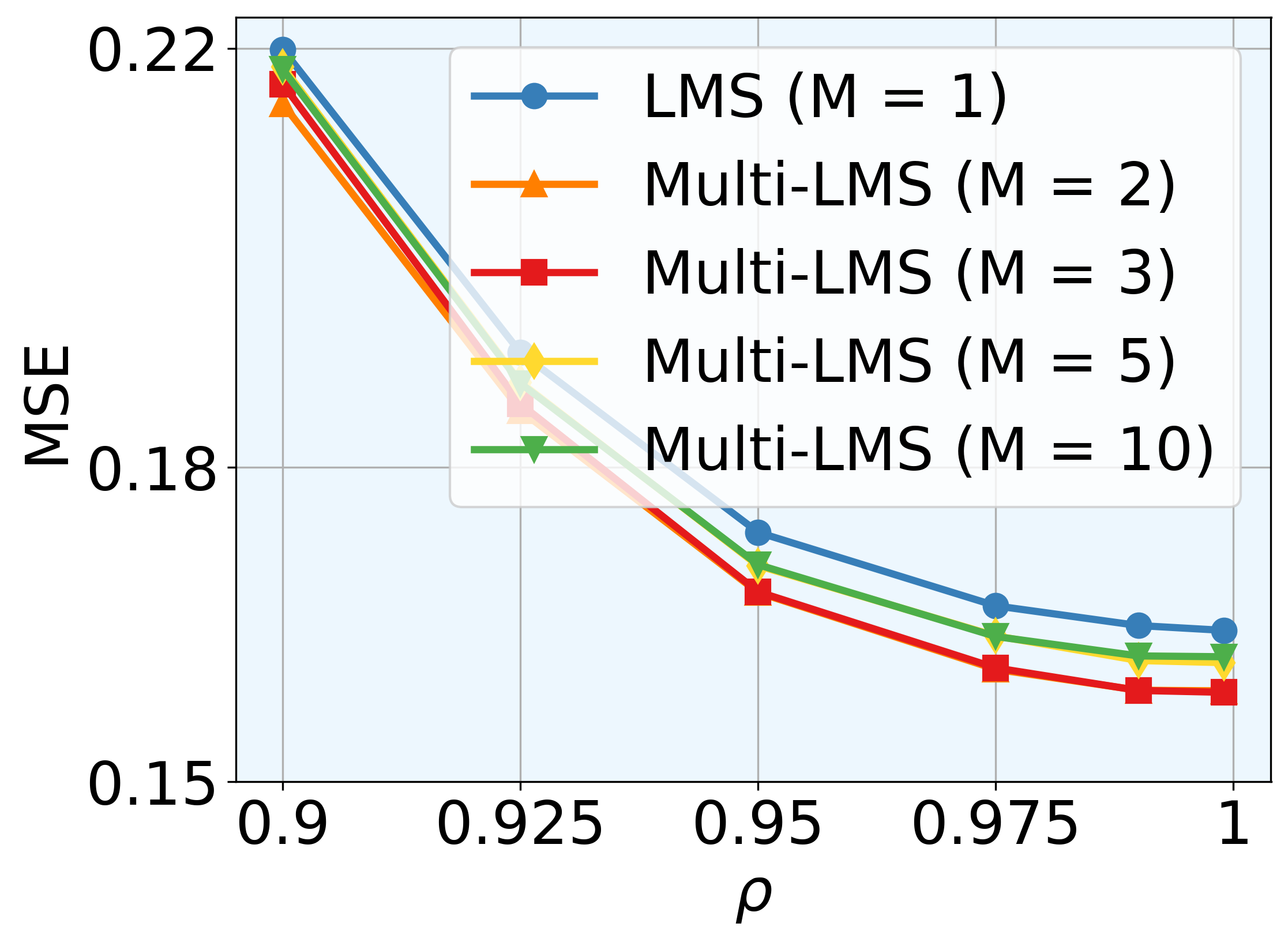}
  }\hfill
  \subfloat[LRMS vs. LMS\label{fig:robust_LRMS}]{
    \includegraphics[width=0.45\linewidth]{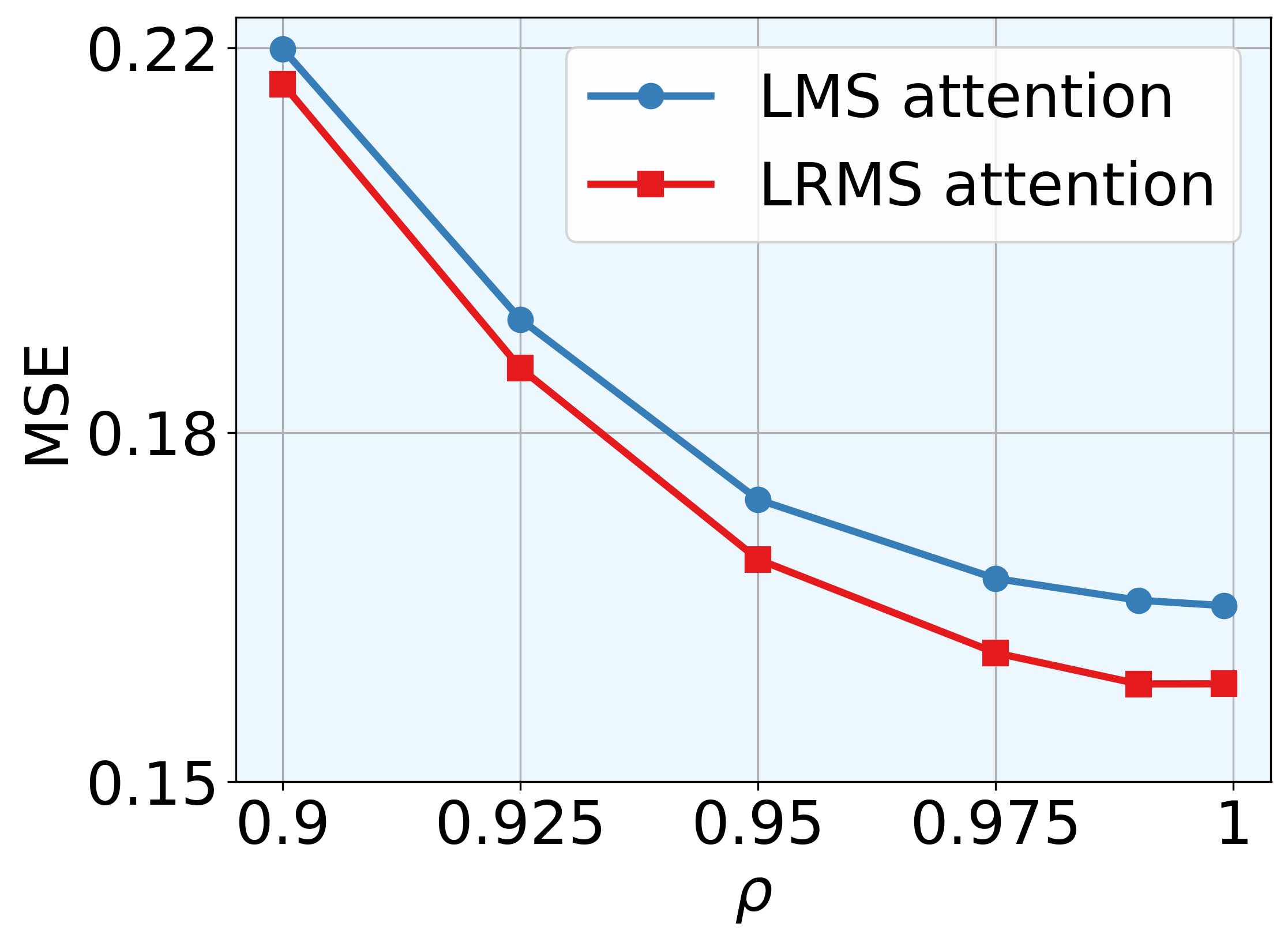}
  }
  \caption{Robustness test with a quantization bit $b = 1$.}
  \vspace{-.1in}
  \label{fig:robustness}
\end{figure}

In the second experiment, we evaluate the effectiveness of Multi-LMS attention, as shown in \Cref{fig:robust_m_step}. We observe that increasing the number of steps $M$ generally improves performance; however, when $M$  becomes too large, the model may overfit to the current noisy observations, resulting in no further improvement and sometimes even degraded performance compared to smaller $M$.

In the last experiment, we compare LMS and LRMS attention, as shown in \Cref{fig:robust_LRMS}. When the quantization bit is low ($b=1$), we observe that LRMS attention outperforms LMS attention. This is because the LRMS mechanism effectively mitigates the impact of outliers and severe quantization noise, resulting in more stable and robust learning.

\section{Conclusion}
\label{sec:conclusion}

In this work, we have investigated the capability of ICL to perform channel equalization in non-stationary MIMO environments. We proposed a principled framework to design attention mechanisms that enhance adaptivity, drawing inspiration from classical adaptive signal processing algorithms. Specifically, we introduced LMS-based attention for rapid adaptation, LRMS formulations to improve robustness against outliers, and multi-step updates to better track long-term channel dynamics. Extensive experiments demonstrated that these adaptive attention mechanisms substantially improve equalization performance in dynamic channels, confirming that ICL can effectively address non-stationary tasks.


\clearpage 

\begingroup
\small            
\setlength{\bibsep}{-0.5pt}      
\setlength{\itemsep}{-0.5pt}     
\setlength{\parsep}{-0.5pt}       
\setlength{\topsep}{-0.5pt}       
\bibliographystyle{IEEEbib}
\bibliography{IEEEabrv,refs}
\endgroup

\end{document}